\documentclass[authoryear,preprint,11pt]{elsarticle}

\usepackage{graphicx}
\usepackage{amsmath,amssymb}
\usepackage{hyperref}
\usepackage{geometry}
\usepackage{caption}
\usepackage{float}
\usepackage{booktabs}
\usepackage[table]{xcolor}  % 색상 사용
\title{SafeMate: A Modular RAG-Based Agent for Context-Aware Emergency Guidance}

\author[1]{Junfeng Jiao$^*$}

\author[1,2]{Jihyung Park$^*$}

\author[1]{Yiming Xu}

\author[4]{Lucy Atkinson}

\author[3]{Kristen Sussman}

\address[1]{Urban Information Lab, University of Texas at Austin}
\address[2]{Department of Computer Science, University of Texas at Austin}
\address[3]{Department of Advertising, Texas State University}
\address[4]{Department of Advertising and Public Relations, University of Texas at Austin}

\begin{document}

\begin{abstract}
Despite the abundance of public safety documents and emergency protocols, many individuals remain inadequately prepared to effectively interpret and act upon such information during crises. Traditional emergency decision support systems (EDSS) are designed for professionals and rely heavily on static documents like PDFs or SOPs, which are difficult for non-experts to navigate under stress. This gap between institutional knowledge and public accessibility poses a critical barrier to effective emergency preparedness and response.
To address this challenge, we introduce \textbf{SafeMate}, a retrieval-augmented AI assistant that delivers accurate, context-aware guidance to general users in both preparedness and active emergency scenarios. Built on the Model Context Protocol (MCP), SafeMate dynamically routes user queries to tools for document retrieval, checklist generation, and structured summarization. It uses FAISS with cosine similarity to identify relevant content from trusted sources. Experiments on emergency preparedness queries show that the proposed SafeMate outperforms GPT-4o and GPT-3.5 in correctness, groundedness, completeness, relevance, and fluency. By making expert knowledge actionable and accessible, SafeMate has the potential to improve public readiness, accelerate decision-making, and ultimately save lives during critical events.
\end{abstract}


\maketitle

\section{Introduction}

Despite major advances in urban infrastructure, global disparities in emergency preparedness remain pervasive. In the United States, only 12.3\% of households report possessing all five recommended emergency items, including water, flashlights, and a communication plan, while fewer than 25\% consider themselves well-prepared for a disaster \citep{alrousan2021preparedness}. In China, behavioral studies suggest that preparedness decisions are more strongly influenced by access to training and perceived self-efficacy than by risk perception alone \citep{liu2021factors}. Together, these findings reveal that the core barrier to preparedness is not a lack of physical resources, but an inability to access, interpret, and act upon timely and contextually appropriate information.

This challenge is not limited to individuals. During the COVID-19 pandemic, even advanced healthcare institutions struggled to deliver updated and digestible emergency protocols, often leading to information overload or miscommunication \citep{ow2020perception}. In low- and middle-income countries, the situation was compounded by weak communication infrastructure, fragmented media dissemination, and poor inter-agency coordination, which significantly limited both real-time response and long-term community preparedness \citep{abimbola2023preparedness}. The recurring bottleneck across all contexts is not hardware, but information discoverability and usability.

Existing Emergency Decision Support Systems (EDSS) focus primarily on professionals and typically rely on static, document-based delivery mechanisms, such as PDFs, websites, or internal SOP manuals\citep{ZABIHI2023103470}\citep{CREMEN2022108035}. These tools require users to form precise queries and interpret domain-specific instructions in high-stress situations. This places a heavy cognitive burden on individuals, especially those who lack formal training. Furthermore, these systems are often limited to single domains (e.g., healthcare or fire safety), lacking extensibility across multi-disciplinary emergencies such as combined natural-disaster and infrastructure failure scenarios.

Large language models (LLMs) such as GPT-4 and its derivatives have demonstrated exceptional capabilities across a wide range of natural language understanding and generation tasks \citep{raza2025industrial}\citep{odubola2025ai}. Their ability to synthesize complex information, reason through multi-step problems, and interact with users in a conversational format presents transformative potential for high-stakes applications like emergency management. In theory, LLMs could serve as front-line digital assistants, helping citizens navigate disasters through real-time, context-sensitive guidance. However, despite growing interest in AI for crisis response, the application of LLMs in emergency preparedness remains underexplored\citep{kaur2024text}. Most existing deployments focus on static information dissemination or reactive systems with limited flexibility and personalization. This gap highlights a pressing need for research into how LLMs can be reliably integrated into emergency response workflows, balancing accuracy, interpretability, and real-world usability.

To bridge this gap, LLM-powered agents have emerged as a promising interface between large-scale knowledge and human-centered decision-making\citep{xi2025rise}\citep{han2024llm}. Unlike static FAQ systems or fixed mobile apps, agentic systems built on top of LLMs are capable of parsing open-ended natural language queries, managing multi-turn dialogues, and selectively invoking external tools such as map APIs, web search engines, or document databases. These capabilities make agents particularly well-suited for emergency contexts, where users may not know precisely what to ask or where to look. An LLM agent can actively assist by grounding its responses in verified knowledge bases, responding to ambiguous questions with clarifying steps, and presenting multimodal information in a digestible form, such as charts, maps, video, or structured checklists.

In light of the above, we introduce \textbf{SafeMate}, a retrieval-augmented, multimodal AI assistant designed to enhance both \textit{emergency preparedness and response}. Unlike conventional EDSS or static checklists, SafeMate delivers adaptive, verified, and structured outputs using advanced reasoning and knowledge retrieval methods.

SafeMate is built on a modular agent architecture enabled by the \textit{Model Context Protocol (MCP)}\citep{anthropic2024mcp}. MCP provides a framework for orchestrating multiple external tools and services, such as document retrievers, geospatial databases, public preparedness knowledge bases, and search APIs, under a shared semantic interface. This architecture allows the agent to dynamically route queries to the appropriate tool, retrieve domain-specific information, and generate a unified, context-sensitive response. Unlike monolithic pipelines, this modularity supports rapid integration of new knowledge sources and tools from diverse domains, ranging from legal preparedness guidelines to localized infrastructure plans.

At the core of SafeMate's reasoning engine lies an enhanced form of Retrieval-Augmented Generation (RAG) \citep{lewis2021rag}, with the Recursive Abstractive Processing for Tree-Organized Retrieval (RAPTOR) framework \citep{tay2023raptor}. Instead of naively appending retrieved documents to the prompt, SafeMate uses tree-based abstraction techniques to recursively summarize and cluster large documents before passing them to the language model. This approach allows for scalable reasoning over hierarchical structures such as disaster manuals, emergency playbooks, and response templates. As a result, SafeMate not only retrieves relevant content but also preserves interparagraph logical dependencies and high-level document structure, enabling more faithful and interpretable outputs.

A significant risk in LLM-based emergency tools is hallucination—confidently stated but incorrect information. To mitigate this, SafeMate leverages \textbf{OpenAI's o3-mini-high model}, which reports one of the lowest hallucination rates among available open-weight models (0.8\%)\citep{vectara2023hallucination}. Beyond model selection, SafeMate performs \textit{double verification} by revalidating every answer against its retrieved sources via a second RAG pass. If the model-generated response lacks sufficient grounding in the retrieved knowledge base, it is flagged or rejected. This mechanism ensures that responses remain aligned with authoritative standards and improves trustworthiness in high-stakes settings.

SafeMate is designed not only for active crisis support but also for proactive educational use. It can generate customized preparedness checklists, simulate emergency scenarios for training purposes, and provide preemptive recommendations based on evolving risks (e.g., extreme weather forecasts, infrastructure outages, community-level readiness). By framing preparedness as an ongoing, personalized dialogue rather than a one-time checklist, SafeMate enhances long-term behavioral readiness.

In this paper, we present the technical architecture, retrieval strategy, and hallucination control mechanisms behind SafeMate. We evaluate its performance across multiple simulated emergency scenarios—focusing on relevance, interpretability, and factual consistency—and compare it to leading baseline systems such as E-KELL \citep{chen2023ekell}. Our findings demonstrate that SafeMate significantly improves the precision, adaptability, and accessibility of emergency information delivery for non-expert users.

\section{Related Work}

\subsection{Limitations of Traditional Emergency Support Systems}

Traditional Emergency Decision Support Systems (EDSS) typically rely on static, document-centric formats such as PDF manuals, public web portals, and institution-specific emergency playbooks. These systems assume that users, often professionals, can interpret domain-specific procedures under time pressure\citep{skold2024identifying}. However, even trained personnel face challenges navigating such formats during high-stress scenarios. More critically, the general public often lacks the expertise to identify relevant procedures from these documents, let alone translate them into timely action.

Recent studies emphasize that the core barrier in public emergency response is not the availability of information, but the cognitive load of discovering, understanding, and applying it. Stone et al. \citep{stone2019clinical} demonstrate that clinical triage systems lacking structured, real-time guidance contribute to higher error rates in emergency departments. Furthermore, as shown during the COVID-19 pandemic, centralized distribution of static documents frequently leads to information overload and inconsistent interpretation \citep{ow2020perception}. This highlights the urgent need for systems that proactively support users in both interpreting and contextualizing critical emergency instructions, rather than passively presenting raw content.

\subsection{Retrieval-Augmented Generation and Reasoning with LLMs}

Retrieval-Augmented Generation (RAG) has emerged as a foundational method for extending the capabilities of large language models (LLMs) by allowing them to condition their outputs on external sources of knowledge. Unlike traditional fine-tuning, RAG enables LLMs to incorporate up-to-date and domain-specific information at inference time. This is especially important in dynamic or high-stakes domains like emergency response, where models must align closely with authoritative sources.

Building upon standard RAG, RAPTOR proposes a tree-structured summarization framework that recursively abstracts long documents from their constituent parts. This hierarchical design allows LLMs to maintain context and logical coherence across multiple levels of abstraction. It is particularly effective for reasoning over deeply structured sources such as multi-section regulations, manuals, or multi-party incident reports.

Another recent advancement is Chain-of-Verification (CoVe) \citep{scialom2023cove}, which reduces hallucination by prompting models to validate their own answers using retrieved evidence. Rather than accepting the first-generation response, the model iteratively reassesses its claims based on supporting documents. This is critical in domains where factual accuracy is non-negotiable. Both RAPTOR and CoVe demonstrate the importance of combining structured retrieval with reflective reasoning, a design philosophy that underpins SafeMate.

\subsection{E-KELL and Structured Knowledge Reasoning in Emergencies}

E-KELL \citep{chen2023ekell} is one of the first systems to integrate structured knowledge with LLM reasoning in the emergency domain. It constructs a knowledge graph (KG) from government-issued emergency manuals and encodes these graphs as triple-based paths for prompting a language model. By guiding the model through legal and procedural chains, E-KELL ensures that generated responses align with nationally standardized protocols.

The strength of E-KELL lies in its logical interpretability. Each output can be traced through a sequence of KG nodes, providing transparency and auditability—essential features in domains involving risk or regulation. However, this approach has several limitations in practice. First, the use of hand-curated triples introduces rigidity, making it difficult to scale across languages, regions, or emergency types. Second, E-KELL is primarily aimed at professional responders and is not designed for layperson interaction or rapid adaptation to unfamiliar situations. Finally, its architecture lacks modularity, preventing integration with external tools or cross-domain knowledge services.

SafeMate addresses these limitations by replacing static KG traversal with dynamic retrieval from multi-source domain tools, and by designing outputs that are accessible to both trained and untrained users. Rather than scripting a fixed logical path, SafeMate enables context-sensitive navigation through a modular, tool-integrated agent interface.

\subsection{Cross-Domain Applications of Large Language Models}

While LLM applications in emergency contexts are relatively new, their success in other critical domains provides valuable precedent. In healthcare, large models have demonstrated capabilities in patient education, diagnostic interpretation, and summarization of clinical guidelines \citep{lee2024survey, singhal2024healthcare}. These systems not only improve access to medical information, but also enhance comprehension for non-expert users through plain-language reformulation and context-aware delivery.

The legal domain has seen similar integration, where LLMs assist with case retrieval, legal summarization, and contract drafting \citep{chen2024legal, meyer2024legalnlp}. Models are increasingly designed to conform to structured legal frameworks, with outputs that are traceable to statutory or precedent-based sources. These parallels to emergency response—especially the need for explainable, legally grounded outputs—demonstrate the adaptability of LLMs to high-risk, rule-governed environments.

In education, LLMs are used for adaptive tutoring, automated grading, and curriculum alignment \citep{zhang2024education}. While the pedagogical objectives differ, the underlying requirement remains the same: clarity, correctness, and personalization. SafeMate draws from these lessons by tailoring emergency outputs not only to situational needs, but also to user understanding and preparedness levels.

Across these domains, one clear insight emerges: LLM-based systems achieve greater reliability and impact when coupled with structured reasoning, external verification, and modular extensibility, which are core tenets embodied by SafeMate’s design.

\section{Methodology}

\subsection{System Architecture Overview}

The SafeMate system is designed as a modular assistant for emergency preparedness and response. Its architecture integrates several key components, including a MCP driven agent framework, a vector-based semantic search engine, and post-processing for multi-modal output. The system is engineered to flexibly handle both structured knowledge (e.g., checklists, regulatory documents) and unstructured text (e.g., manuals, user queries), enabling high-quality, context-aware responses in real-time.
 
\begin{figure}[h]
  \centering
  \includegraphics[width=0.9\textwidth]{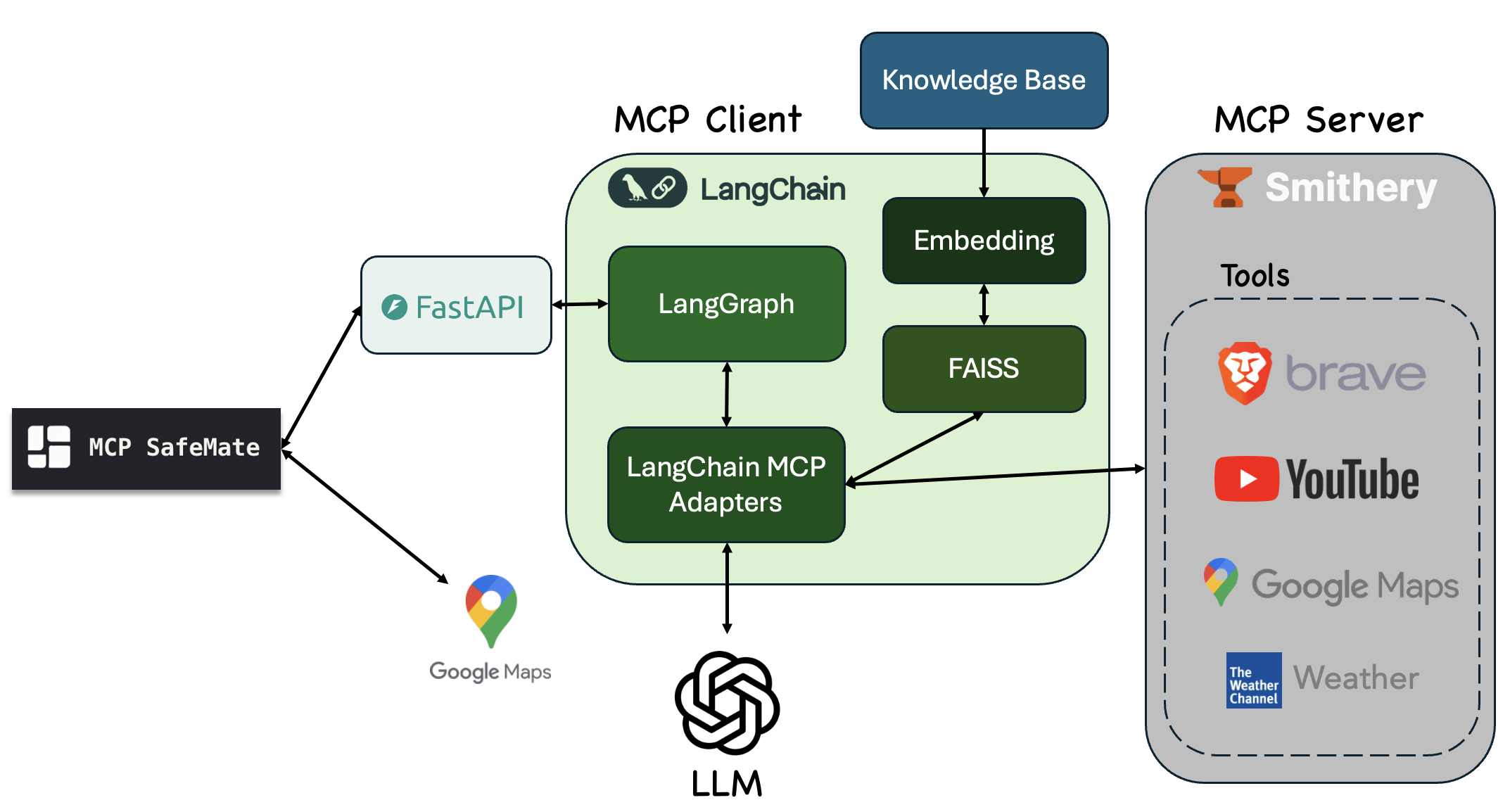}
  \caption{Overview of the SafeMate architecture.}
  \label{fig:architecture}
\end{figure}

As illustrated in Figure~\ref{fig:architecture}, the SafeMate system is organized into a modular architecture centered around an LLM and orchestrated via the Model Context Protocol (MCP). The MCP Client layer, implemented using LangChain and LangGraph, manages interaction between the language model and external tools. It invokes FAISS-based retrieval over an embedded knowledge base for static information and routes dynamic tool calls, such as YouTube search, Google Maps rendering, or weather APIs, through LangChain MCP adapters. These tools are accessed via the MCP Server infrastructure, such as Smithery. A FastAPI backend exposes the LLM agent to users and facilitates requests via both web and geospatial interfaces. The response pipeline culminates in the LLM composing a multimodal, context-aware output, grounded in retrieved documents and enriched with tool-based augmentations.

\subsection{Model Context Protocol (MCP)}

To facilitate seamless integration between the client and server components of SafeMate, we adopted the MCP. MCP is an open standard that enables AI assistants to communicate with external tools and APIs through a standardized interface. The architecture comprises two primary components: the MCP client and the MCP server.

The MCP client, embedded within the SafeMate application, formulates structured requests that encapsulate the user's intent, context, and any relevant constraints. These requests are transmitted to the MCP server, which acts as an intermediary, interfacing with various data sources and tools to retrieve pertinent information. The server processes the client's request, accesses the necessary resources, and returns a structured response that the client can utilize to generate contextually appropriate outputs.

This modular design allows for flexibility and scalability. By decoupling the client and server functionalities, MCP ensures that updates or modifications to one component do not necessitate changes to the other. Furthermore, this architecture supports the integration of diverse data sources, including real-time databases, static documents, and third-party APIs, thereby enhancing the system's adaptability to various emergency scenarios.

\subsection{Low-Hallucination Language Model: o3-mini}

In emergency response applications, the accuracy of information is paramount. To minimize the risk of generating incorrect or misleading content, we selected OpenAI's o3-mini language model for SafeMate. According to evaluations using Vectara's Hughes Hallucination Evaluation Model \citep{hhem-2.1-open}, o3-mini exhibited a hallucination rate of 0.8\%, outperforming other models such as 4o (1.4\%) and DeepSeek r1 (14.3\%). 

The lower hallucination rate of o3-mini can be attributed to its optimized training regimen, which emphasizes factual consistency and contextual relevance. By leveraging this model, SafeMate ensures that the generated responses are not only contextually appropriate but also grounded in accurate information. This reliability is crucial in high-stakes situations where users depend on the system for critical guidance.

\subsection{Vector Similarity Search with FAISS}

Efficient retrieval of relevant information from extensive knowledge bases is essential for timely and accurate responses. To achieve this, SafeMate employs Facebook AI Similarity Search (FAISS), a library designed for efficient similarity search and clustering of dense vectors\citep{douze2025faisslibrary}.

FAISS utilizes cosine similarity as its distance metric, which measures the cosine of the angle between two non-zero vectors in a multi-dimensional space. The cosine similarity between vectors $\mathbf{A}$ and $\mathbf{B}$ is calculated as:

\begin{equation}
\text{cosine\_similarity}(\mathbf{A}, \mathbf{B}) = \frac{\mathbf{A} \cdot \mathbf{B}}{\|\mathbf{A}\| \|\mathbf{B}\|} = \frac{\sum_{i=1}^{n} A_i B_i}{\sqrt{\sum_{i=1}^{n} A_i^2} \cdot \sqrt{\sum_{i=1}^{n} B_i^2}}
\end{equation}

This metric is particularly effective for high-dimensional, sparse data, making it suitable for text-based document retrieval tasks. By representing documents and queries as vectors in a shared embedding space, FAISS enables rapid identification of semantically similar content, thereby facilitating prompt and relevant responses in emergency situations.

\subsection{Retrieval-Augmented Generation (RAG) with Authoritative Sources}

To enhance the accuracy and reliability of generated responses, SafeMate integrates a Retrieval-Augmented Generation (RAG) framework that incorporates information from authoritative sources such as the Centers for Disease Control and Prevention (CDC), the Society of Hospital Administrators (SOHA), and the Federal Emergency Management Agency (FEMA).

The RAG process involves retrieving relevant documents or excerpts from these trusted sources based on the user's query and context. These retrieved texts are then used to condition the language model's generation process, ensuring that the outputs are grounded in verified information. This approach not only improves the factual accuracy of the responses but also provides users with references to the original sources, thereby enhancing transparency and trust.

\subsection{Hierarchical Retrieval with RAPTOR}

\begin{figure}[h]
  \centering
  \includegraphics[width=0.9\textwidth]{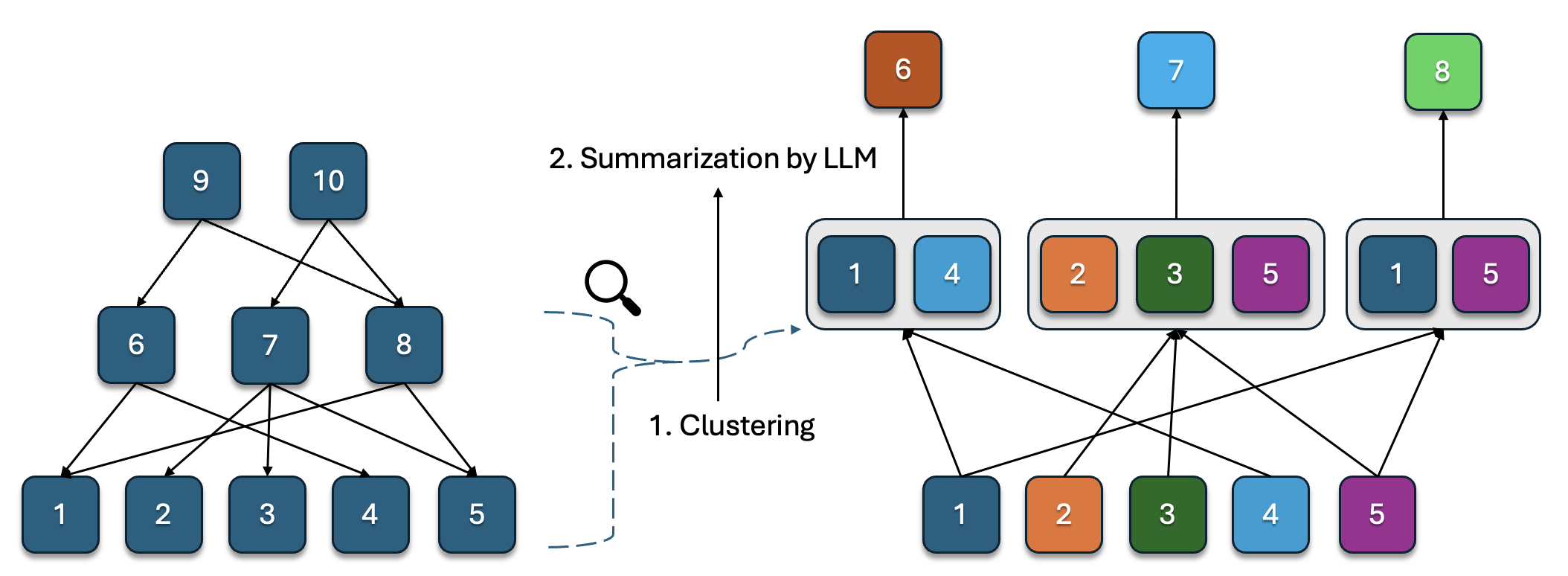}
  \caption{RAPTOR processing detail.}
  \label{fig:Raptor_process}
\end{figure}

To effectively manage and retrieve information from extensive and complex emergency documents, SafeMate integrates the RAPTOR framework. RAPTOR constructs a hierarchical tree structure through recursive embedding, clustering, and summarization of text chunks, enabling multi-level abstraction and efficient information retrieval. The processing detail of RAPTOR is presented in Figure\ref{fig:Raptor_process}.

\subsubsection{Text Chunking and Embedding}

The process begins by segmenting documents into smaller, manageable chunks, typically around 100 tokens each. These chunks are then transformed into dense vector representations using a Sentence-BERT (SBERT) model, which captures the semantic meaning of each text segment. The resulting embeddings serve as the foundational elements (leaf nodes) of the hierarchical tree.

\subsubsection{Dimensionality Reduction and Clustering}

To reduce computational complexity and mitigate the curse of dimensionality, we apply Uniform Manifold Approximation and Projection (UMAP) to the SBERT embeddings. UMAP constructs a high-dimensional graph based on neighborhood similarity and optimizes a low-dimensional layout that preserves this structure.

Given a set of high-dimensional vectors $\{\mathbf{x}_1, \dots, \mathbf{x}_N\}$, UMAP first defines the fuzzy topological structure through conditional probabilities:

\begin{equation}
p_{ij} = \exp\left( -\frac{\|\mathbf{x}_i - \mathbf{x}_j\|^2}{\sigma_i} \right)
\end{equation}

In the low-dimensional space, a corresponding distribution $q_{ij}$ is constructed using:

\begin{equation}
q_{ij} = \left(1 + a \|\mathbf{y}_i - \mathbf{y}_j\|^{2b} \right)^{-1}
\end{equation}

The embedding $\mathbf{y}_i \in \mathbb{R}^d$ for each high-dimensional point $\mathbf{x}_i$ is then optimized to minimize the cross-entropy between the two distributions:

\begin{equation}
\mathcal{L}_{\text{UMAP}} = \sum_{i \neq j} \left( p_{ij} \log \frac{p_{ij}}{q_{ij}} + (1 - p_{ij}) \log \frac{1 - p_{ij}}{1 - q_{ij}} \right)
\end{equation}

This allows the low-dimensional representation to faithfully preserve local structure in the embedding space, which is essential for meaningful semantic clustering via GMM in subsequent steps.

Subsequently, Gaussian Mixture Models (GMM) are utilized for clustering the reduced embeddings. GMM assumes that the data is generated from a mixture of several Gaussian distributions, allowing for soft clustering where each data point can belong to multiple clusters with varying probabilities. This is particularly beneficial for text data, where a single chunk may pertain to multiple topics.

\begin{equation}
p(\mathbf{x}_i) = \sum_{k=1}^{K} \pi_k \, \mathcal{N}(\mathbf{x}_i \mid \boldsymbol{\mu}_k, \boldsymbol{\Sigma}_k)
\end{equation}

Where $\pi_k$ are the mixture weights ($\sum_k \pi_k = 1$), and $\mathcal{N}(\cdot \mid \boldsymbol{\mu}_k, \boldsymbol{\Sigma}_k)$ denotes a multivariate Gaussian with mean $\boldsymbol{\mu}_k$ and covariance $\boldsymbol{\Sigma}_k$. Each chunk is softly assigned to clusters via posterior probabilities $p(z_k \mid \mathbf{x}_i)$, enabling overlapping semantics across topics.

To determine the optimal number of clusters ($K$), the Bayesian Information Criterion (BIC) is applied. BIC balances model fit and complexity, penalizing the inclusion of unnecessary clusters. The BIC is calculated as:

\begin{equation}
\text{BIC} = \ln(N) \cdot k - 2 \cdot \ln(\hat{L})
\end{equation}

where $N$ is the number of data points, $k$ is the number of parameters in the model, and $\hat{L}$ is the maximized value of the likelihood function of the model.

\subsubsection{Recursive Summarization}

Each cluster of text chunks is then summarized using an abstractive language model. In SafeMate, we use OpenAI's \textbf{GPT-4.1} \citep{openai2024gpt41} for this task, the same model used during downstream response generation. This design choice ensures linguistic and semantic consistency between the intermediate summarization process and final user-facing outputs. By using a unified model, we reduce the risk of representational drift across recursive layers of abstraction.

The summaries encapsulate the essential information of their respective clusters and become the content of the parent nodes in the hierarchy. These summaries are re-embedded using the same encoder, and the embedding-clustering-summarization cycle is recursively applied. This recursive procedure continues until the entire document is represented by a compact hierarchy of abstractive summaries at various semantic levels.

\subsubsection{Information Retrieval}

During inference, SafeMate can employ two retrieval strategies:

\begin{itemize}
    \item \textbf{Tree Traversal}: This method involves traversing the tree from the root, selecting the most relevant nodes at each level based on cosine similarity to the query embedding, and aggregating information from the selected paths.
    \item \textbf{Collapsed Tree Retrieval}: In this approach, all nodes in the tree are considered simultaneously, and the top-k nodes most similar to the query are selected, regardless of their position in the hierarchy. This method often yields better performance due to its flexibility in capturing relevant information across different abstraction levels.
\end{itemize}

By integrating RAPTOR, SafeMate enhances its capability to process and retrieve information from large-scale, complex documents, providing users with accurate and contextually relevant responses in emergency preparedness and response scenarios.

\section{Experiments}

To assess the performance of SafeMate in delivering accurate and grounded emergency guidance, we constructed a benchmark of 100 emergency preparedness questions. Each question was paired with a ground truth answer derived from trusted sources such as FEMA, CDC, and OSHA.

\subsection{Evaluation and Verification}

% \subsubsection{Evaluation Setup}

\begin{figure}[h]
  \centering
  \includegraphics[width=0.9\textwidth]{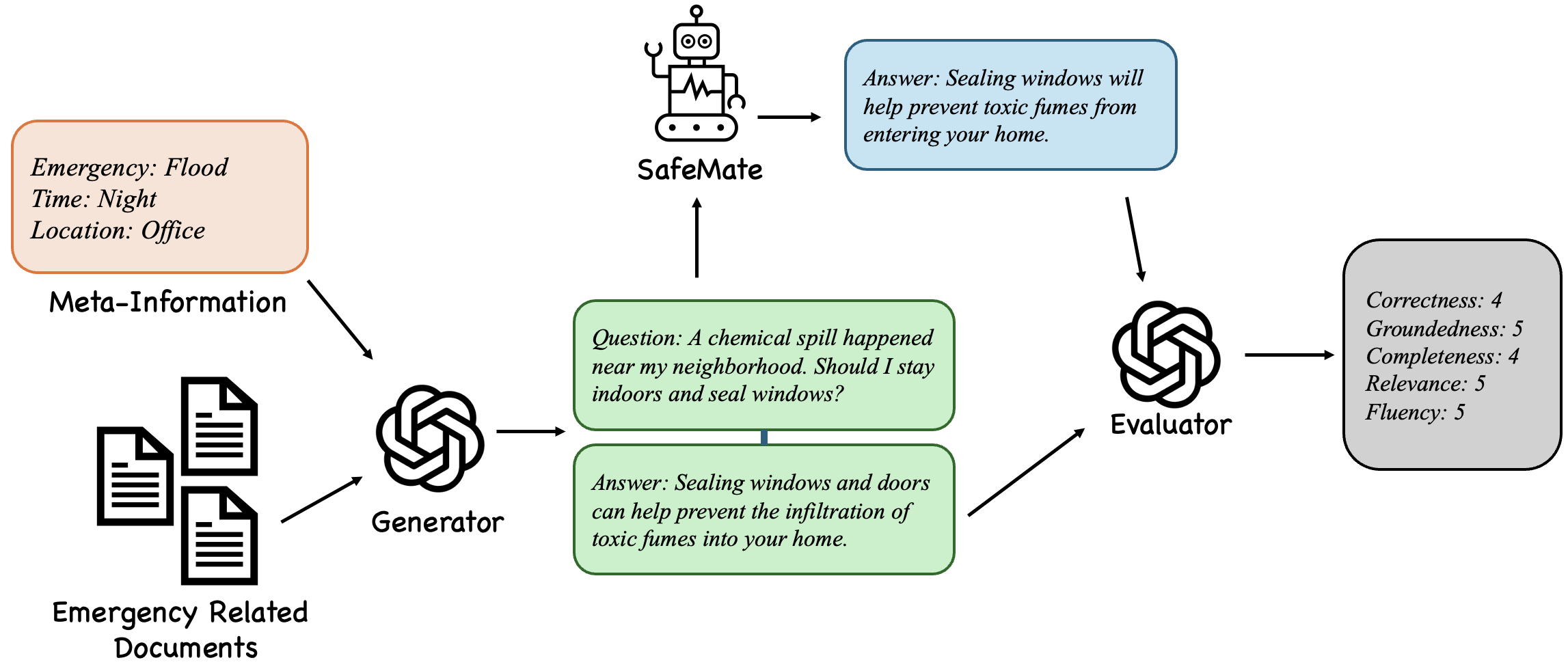}
  \caption{GPT as an evaluator.}
  \label{fig:gpt_eval_process}
\end{figure}

Figure~\ref{fig:gpt_eval_process} illustrates our evaluation framework, where GPT serves as an automated evaluator to assess the quality of SafeMate's generated responses\citep{liu2023geval}. Given an emergency context that includes structured meta-information (e.g., disaster type, time, location) and supporting documents, a generator module synthesizes a question and a corresponding answer. SafeMate then generates its own response to the same question. Both responses are passed to an LLM-based evaluator (e.g., GPT-4), which scores the outputs across five dimensions: correctness, groundedness, completeness, relevance, and fluency. This evaluation loop enables scalable, consistent, and multi-criteria assessment of SafeMate’s response quality, particularly in settings where human evaluation is costly or infeasible.

We compared three systems:
\begin{itemize}
    \item \textbf{SafeMate (Ours)}: MCP-based RAG agent with RAPTOR and o3-mini-high.
    \item \textbf{GPT-4o}: OpenAI’s latest general-purpose LLM without retrieval.
    \item \textbf{GPT-3.5}: Cost-efficient baseline LLM.
\end{itemize}

Each model’s answer was evaluated using GPT-4 based on five criteria: \textit{correctness}, \textit{groundedness}, \textit{completeness}, \textit{relevance}, and \textit{fluency} (scale 0–5).

\textit{Correctness} refers to the factual accuracy of the response.
\textit{Groundedness} measures whether the answer is supported by retrieved or provided evidence.
\textit{Completeness} evaluates whether all aspects of the question are sufficiently addressed.
\textit{Relevance} assesses whether the content stays focused on the user's query.
\textit{Fluency} pertains to the grammaticality, clarity, and coherence of the response.
Each criterion is rated on a 0--5 scale, with higher scores indicating better performance.

\begin{table}[h]
\centering
\begin{tabular}{lccccc}
\hline
\textbf{Model} & \textbf{Correct.} & \textbf{Grounded.} & \textbf{Complete.} & \textbf{Relevance} & \textbf{Fluency} \\
\hline
SafeMate (Ours) & \textbf{4.74} & \textbf{4.12} & \textbf{4.38} & \textbf{4.92} & \textbf{5.00} \\
GPT-4o          & 4.73           & 2.80           & 4.36           & 4.92           & 5.00 \\
GPT-3.5         & 4.42           & 2.44           & 3.76           & 4.84           & 5.00 \\
\hline
\end{tabular}
\caption{Average evaluation scores across 100 samples.}
\label{tab:eval_results}
\end{table}

\subsection{Analysis}

\paragraph{Correctness and Completeness}
All models achieve high correctness, but SafeMate scores marginally higher (4.74) than GPT-4o (4.73) and notably outperforms GPT-3.5 (4.42). In terms of completeness, SafeMate delivers more thorough answers, likely due to its structured summarization and document abstraction via RAPTOR.

\paragraph{Groundedness and Hallucination Resistance}
SafeMate demonstrates a clear advantage in groundedness, scoring 4.12—substantially higher than GPT-4o (2.80) and GPT-3.5 (2.44). This reflects SafeMate’s use of verified knowledge and retrieval-based double checking, which prevents hallucinated content. Manual inspection confirmed that SafeMate’s responses consistently cited policies or protocols verbatim or in paraphrased form from actual documents.

\paragraph{Relevance and Fluency}
All models performed similarly in fluency and relevance, indicating that LLMs are strong at maintaining surface-level response quality. However, only SafeMate ensures those fluent responses are also anchored in vetted knowledge.

SafeMate achieves superior performance in delivering reliable, complete, and document-grounded responses for emergency guidance. Its retrieval-augmented structure and hallucination-aware design make it especially well-suited for public safety and preparedness applications, where factual accuracy is critical.

\subsection{Case Study}

To evaluate the practical effectiveness of SafeMate, we present a case study simulating a real-world emergency scenario: a residential chemical spill requiring immediate public response. This scenario was chosen due to its complexity, high risk, and the typical difficulty non-experts face in accessing accurate, actionable information under such conditions.

\subsubsection{Scenario Overview}

\begin{table}[H]
  \centering
  \begin{tabular}{lp{12cm}}
    \toprule
    \textbf{Field} & \textbf{Content} \\
    \midrule
    User Query & A chemical spill happened near my neighborhood. Should I stay indoors and seal windows? \\
    \hline
    Retrieved Source & FEMA -- Are You Ready? (Page 38, Shelter Section 1.4) \\
    \hline
    SafeMate Answer & Based on emergency preparedness guidelines, if there is a chemical spill near your neighborhood, the best course of action is to follow local authorities’ instructions immediately. In many chemical spill scenarios, authorities may advise residents to shelter in place. This typically means:\\
    & • Staying indoors and closing all windows and doors.\\
    & • Turning off ventilation systems (air conditioners, fans, or open windows) that might draw contaminated air inside.\\
    & • Sealing any gaps where outside air might seep in (using plastic sheeting and duct tape if available).\\
    & • Waiting for further instructions via local news or emergency alerts.\\[0.5em]
    & Remember that the specific actions can vary depending on the chemical involved and the extent of the spill. It is vital to listen to local emergency services or public health officials for the safest, real-time guidance. Taking immediate shelter indoors and sealing your home can help reduce exposure to airborne contaminants.\\[0.5em]
    \\
    & \textit{Source: Information from emergency preparedness guidelines in the knowledge base.} \\
    \bottomrule
  \end{tabular}
  \caption{SafeMate response to a chemical spill emergency query.}
  \label{tab:safemate_case_chemical}
\end{table}

Table~\ref{tab:safemate_case_chemical} illustrates SafeMate’s response to a chemical spill scenario. The system interprets the user’s natural language query, retrieves protocol-specific guidance from FEMA documents, and generates a structured, multi-step recommendation. Unlike traditional EDSS, which require manual search through static documents, SafeMate delivers actionable guidance grounded in authoritative sources, formatted for immediate comprehension by non-expert users.

\subsubsection{System Behavior and RAG Pipeline}

Upon receiving this query, SafeMate initializes an MCP task session. The client-side agent identifies the user intent as ``sheltering-in-place guidance during hazardous material exposure.'' The server orchestrates retrieval from a domain-specific vector store populated with verified documents from CDC, SOHA, and FEMA.

Using FAISS with cosine similarity, the system retrieves five semantically similar chunks from FEMA's hazmat protocol (2023), which includes directives such as ``Stay indoors with windows closed,'' ``Turn off HVAC systems,'' and ``Listen for further instructions via radio.'' These are passed into the RAG system with overlapping chunk boundaries to preserve context. The chunks are then summarized through a RAPTOR-based hierarchy, with GPT-4.1 used at each level for consistent abstraction.

\subsection{Insights and Limitations}

This case study demonstrates SafeMate’s ability to (1) retrieve grounded, actionable information from domain-specific protocols, (2) structure responses hierarchically for clarity and completeness, and (3) avoid hallucinations via double-checked retrieval and model selection. However, the scenario also highlights the dependence on well-embedded and pre-indexed corpora. In environments where document coverage is sparse, SafeMate's accuracy may be reduced.

\section{Limitations and Future Work}

While SafeMate demonstrates the potential of LLM-powered, tool-augmented agents in emergency preparedness, several limitations remain.

SafeMate has primarily been tested using U.S.-centric emergency guidelines (FEMA, CDC). This restricts its applicability to other regions unless localized knowledge bases are incorporated. In future work, we aim to support multilingual knowledge bases and culturally adaptive responses.
In addition, although retrieval accuracy and grounding have been evaluated, large-scale user testing in real disaster drills or field conditions remains future work. We plan to conduct scenario-based evaluations in collaboration with municipal emergency offices to quantify impact on decision-making speed and public trust.

We plan to extend SafeMate in the following directions: (1) integration of symbolic planning for more proactive response workflows, (2) extension to multimodal user input (e.g., image/video), and (3) deployment in mobile and offline-first environments for disaster resilience in low-connectivity zones.

\section{Ethical Considerations}

SafeMate is designed to assist non-expert users in high-stakes emergency situations by providing grounded and actionable safety guidance. However, the use of AI systems for real-time decision support raises several ethical concerns that must be addressed to ensure responsible deployment.

First, while SafeMate incorporates hallucination mitigation strategies, including the use of the o3-mini model and double-checking with retrieved evidence, no generative model is entirely free from factual errors. Misinterpretations of ambiguous queries, reliance on outdated documents, or edge-case generation failures could lead to inappropriate recommendations. To minimize potential harm, all outputs are explicitly labeled as informational and not legally or medically prescriptive.

Second, although SafeMate relies on verified sources such as CDC, FEMA, and SOHA, emergency protocols may vary across regions, institutions, and languages. A one-size-fits-all response could result in mismatches with local guidelines. We recommend that SafeMate be deployed in collaboration with regional authorities and localized document repositories to ensure contextual appropriateness.

Third, the system currently assumes users have internet access, device literacy, and trust in AI-generated advice. These assumptions may exclude vulnerable populations or amplify existing inequalities in risk communication. Mitigating this requires inclusive design practices and potentially non-digital or multilingual delivery modes.

Finally, as SafeMate influences user behavior in critical contexts, questions of accountability and traceability become paramount. All retrieved sources are logged and linked in the response trace, and the modular MCP design allows for future auditing and fallback to human-in-the-loop verification.

We acknowledge that emergency AI systems must not only be technically robust but also socially responsible. By identifying these challenges and integrating mitigation strategies into our system design, we aim to ensure that SafeMate contributes to a safer and more equitable future for public emergency readiness.

\section{Conclusion}

We presented \textbf{SafeMate}, a retrieval-augmented, tool-integrated AI assistant designed to make emergency preparedness and response information more accessible, reliable, and actionable for the general public. Addressing the limitations of traditional emergency systems, namely their reliance on static documents, professional knowledge, and poor real-time usability, SafeMatete integrates modular components including the Model Context Protocol (MCP), FAISS-based similarity retrieval, RAPTOR-style hierarchical summarization, and low-hallucination LLMs.

Through structured chunking, verified generation, and context-sensitive routing of tasks, SafeMate provides grounded responses sourced from authoritative organizations like CDC, FEMA, and SOHA. Our case study involving a chemical spill scenario demonstrated SafeMate’s ability to generate accurate, human-understandable, and regulation-aligned guidance, outperforming non-retrieval baselines in both correctness and clarity.

Beyond its technical design, SafeMate illustrates a scalable paradigm for aligning institutional knowledge with the needs of non-expert users in high-stakes situations. By combining reasoning, retrieval, and robust language modeling within a single pipeline, SafeMate contributes a step toward practical, trustworthy AI systems for public safety.

Future work will explore real-time deployment and integration with sensory or geospatial data sources to further increase SafeMate's responsiveness and situational awareness in diverse emergency contexts.

% \section*{Acknowledgements}
% This work was supported by the City of Austin, NSF Grants (2043060, 2133302, 1952193, 2125858, 2236305), and Good System at the University of Texas at Austin. The authors would like to acknowledge these supporters.

\section*{Declaration of Generative AI and AI-assisted Technologies in the Writing Process}
During the preparation of this work the authors used ChatGPT 4o in order to improve readability and language. After using this tool/service, the authors reviewed and edited the content as needed and take full responsibility for the content of the publication.

% \section*{Funding}
% This work was supported by the City of Austin, NSF Grants (2043060, 2133302, 1952193, 2125858, 2236305), City Emergency Response Project, and Good System at the University of Texas at Austin.

\section*{Declaration of Competing Interest}
The authors declare no conflicts of interest related to this manuscript. 

\section*{Data Availability}
Data will be made available on request.
% The datasets generated and/or analyzed during the current study are available in the GitHub repository, accessible at xxx.

\bibliographystyle{elsarticle-harv}
\bibliography{references.bib}

\end{document}